%% file: main.tex
\documentclass[10pt,twocolumn,letterpaper]{article}

\usepackage[pagenumbers]{cvpr} 

\input{preamble}

\definecolor{cvprblue}{rgb}{0.21,0.49,0.74}
\usepackage[pagebackref,breaklinks,colorlinks,allcolors=cvprblue]{hyperref}

\def\paperID{*****} 
\def\confName{CVPR}
\def\confYear{2026}

\title{Q-DeepSight: Incentivizing Thinking with Images for Image Quality \\ Assessment and Refinement}

\author{
Xudong Li$^{1}$ \quad
Jiaxi Tan$^{1}$ \quad
Ziyin Zhou$^{1}$ \quad
Yan Zhong$^{2}$ \quad
Zihao Huang$^{3}$ \\
Jingyuan Zheng$^{1}$ \quad
Yan Zhang$^{1}$ \quad
Xiawu Zheng$^{1}$ \quad
Rongrong Ji$^{1}$ \\[6pt]
$^{1}$\,Key Laboratory of Multimedia Trusted Perception and Efficient Computing,\\
Ministry of Education of China, Xiamen University, 361005, P.R. China \\
$^{2}$\,Peking University \qquad
$^{3}$\,Beijing Institute of Technology
}

\begin{document}

\twocolumn[{
\renewcommand\twocolumn[1][]{#1}
\maketitle
\centering
\vspace{-0.4cm}
\includegraphics[width=\textwidth]{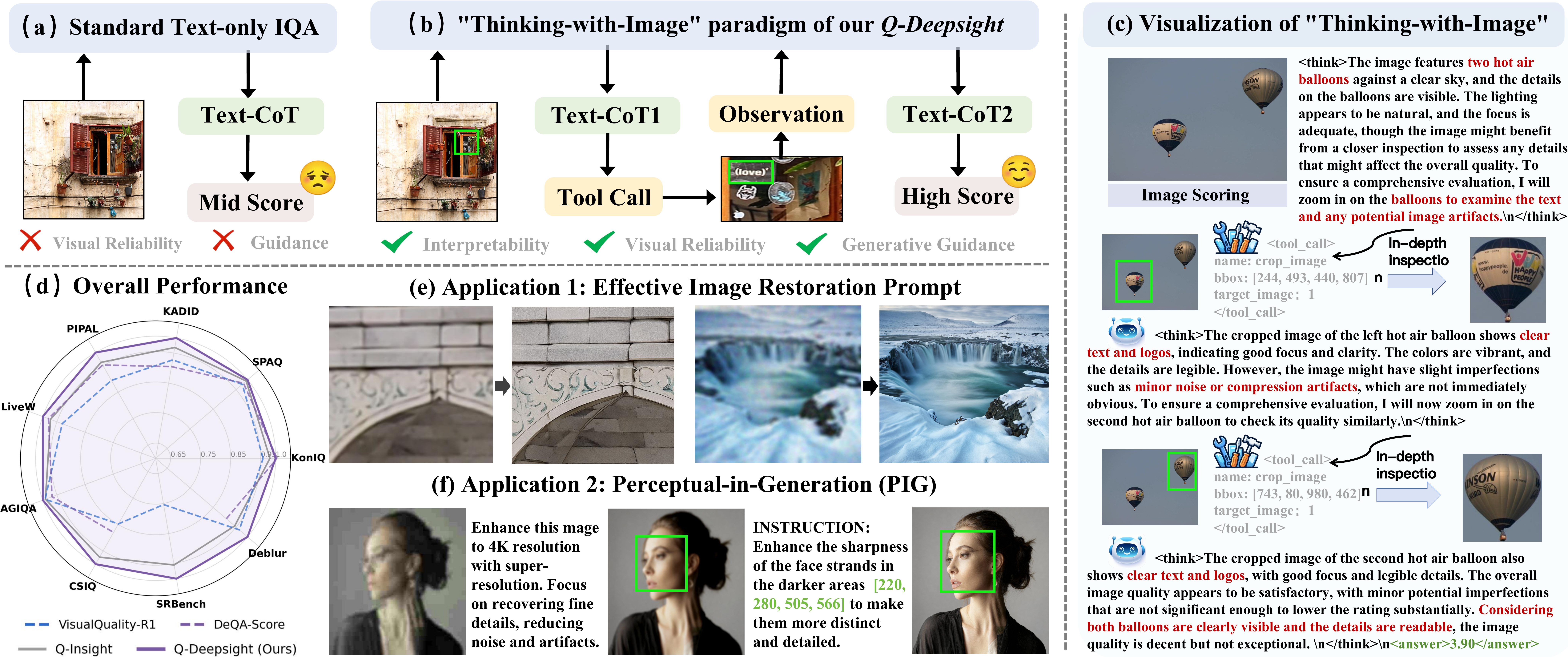}
\vspace{-0.5cm}
\captionsetup{type=figure}
\caption{\textbf{Overview of Q-DeepSight and its applications.} (a) Standard text-only IQA lacks visual grounding, leading to unreliable assessments. (b) Our \textit{``Thinking-with-Image''} paradigm interleaves reasoning with tool-augmented observations for enhanced interpretability and reliability. (c) Visualization showing hierarchical zoom-in operations for fine-grained detail verification. (d) Performance comparison across 9 datasets demonstrating Q-DeepSight's superiority. (e-f) Our method empowers downstream applications, including generating prompts for image restoration and perceptual-guided iterative image enhancement.}
\label{图1}
}
\vspace{0.6cm}]

\input{sec/0_abstract}

\input{sec/1_intro}
\input{sec/2_related_work}

\input{sec/3_methodology}
\input{sec/4_experiments}
\input{sec/5_conclusion}

{
    \small
    \bibliographystyle{ieeenat_fullname}
    \bibliography{sample-base}
}

\end{document}

%% file: preamble.tex
\usepackage{subcaption}
\usepackage{bbm}
\usepackage{algorithmic}
\usepackage{multirow}
\usepackage{soul}
\usepackage{anyfontsize}
\usepackage{mathtools}
\usepackage{colortbl}
\usepackage{wrapfig}
\usepackage{enumitem}
\usepackage{tcolorbox}

\definecolor{customgreen}{rgb}{0.0, 0.7, 0.3}

\providecommand{\Description}[1]{}

%% file: sec/0_abstract.tex
\begin{abstract}
Image Quality Assessment (IQA) models are increasingly deployed as perceptual critics to guide generative models and image restoration. This role demands not only accurate scores but also actionable, localized feedback.
However, current MLLM-based methods adopt a \emph{single-look, language-only} paradigm, which departs from human evidence-seeking judgment and yields weakly grounded rationales, limiting their reliability for in-the-loop refinement.
We propose \emph{\textbf{Q-DeepSight}}, a \emph{think-with-image} framework that emulates this human-like process. It performs {interleaved Multimodal Chain-of-Thought (iMCoT)} with tool-augmented evidence acquisition (e.g., crop-and-zoom) to explicitly determine \emph{where} quality degrades and \emph{why}.
To train these long iMCoT trajectories via reinforcement learning, we introduce two techniques: {Perceptual Curriculum Reward (PCR)} to mitigate reward sparsity and {Evidence Gradient Filtering (EGF)} to improve credit assignment for visually-grounded reasoning.
Q-DeepSight achieves state-of-the-art performance across diverse benchmarks, including natural, restored, and AI-generated content.
Furthermore, we demonstrate its practical value with {Perceptual-in-Generation (PiG)}, a training-free framework where Q-DeepSight's diagnoses guide iterative image enhancement, effectively closing the loop between assessment and refinement. Our code will be released soon.
\end{abstract}

%% file: sec/1_intro.tex
\section{Introduction}
Image Quality Assessment (IQA) aims to quantify perceptual quality in agreement with human judgment \cite{reiqa,NIQE,ke2021musiq}. Its importance has grown rapidly as IQA models are increasingly used as \emph{perceptual critics} in modern vision pipelines, serving as preference signals to align generative models \cite{wang2025unified} and as optimization targets for image restoration \cite{he2024videoscore}. This shift requires modern IQA to not only generalize across natural and AI-generated content, but also be \emph{explainable} and \emph{actionable}, providing grounded evidence of \emph{where} and \emph{why} quality degrades. 
Recent progress in Multimodal Large Language Models (MLLMs) makes this feasible by producing both scores and natural-language critiques. Current methods typically fall into two lines: \textbf{score-centric} methods that emphasize score accuracy but lack grounding \cite{you2025teaching,wu2023q}, and \textbf{description-centric} methods that generate rationales but rely heavily on supervised fine-tuning (SFT) \cite{descriptqa}. While recent works adopt reinforcement learning (RL) to reduce this annotation dependency \cite{li2025q,wu2025visualquality}, these models still exhibit two key limitations.

\textbf{First, it departs from human perceptual judgment.}
Existing methods follow a \emph{single-look, language-only} paradigm: the image is encoded once into a global representation and the model then ``justifies'' the score largely in language space, yielding superficially plausible rationales that are often weakly supported by concrete visual evidence. In contrast, human quality judgment is \emph{active} and evidence-seeking, involving iterative inspection and zooming into candidate regions to verify localized artifacts.
\textbf{Second, this weak grounding makes them unreliable} as perceptual critics for image generation or restoration. A practical critic should provide \emph{actionable} and \emph{fine-grained} feedback---grounded in \emph{where} quality degrades and \emph{why}---so downstream systems can decide what to refine, especially for high-resolution synthesis and region-dependent artifacts (e.g., ringing, blockiness, texture loss, and over-smoothing). These observations suggest that turning IQA into a reliable critic requires models to \emph{actively acquire} and revisit targeted visual evidence during reasoning, enabling grounded and actionable assessments.

To close this gap, we propose \emph{\textbf{Q-DeepSight}}, a \emph{think-with-image} IQA framework that performs {interleaved Multimodal Chain-of-Thought (iMCoT)} with tool-augmented evidence acquisition (e.g., bbox-guided crop-and-zoom).
As shown in Fig.~\ref{图1}, instead of relying on a single global encoding, Q-DeepSight alternates between (i) generating a textual reasoning step based on the current visual evidence, and (ii) actively querying for new visual observations by using tools to inspect specific distortion regions, thereby explicitly learning \emph{where} to look and \emph{what} to conclude. This reformulates IQA as a sequential decision process that yields region-level, evidence-backed justifications, making it a more reliable critic for downstream refinement and decision-making.

\begin{figure}[t]
  \centering
  \includegraphics[width=\columnwidth]{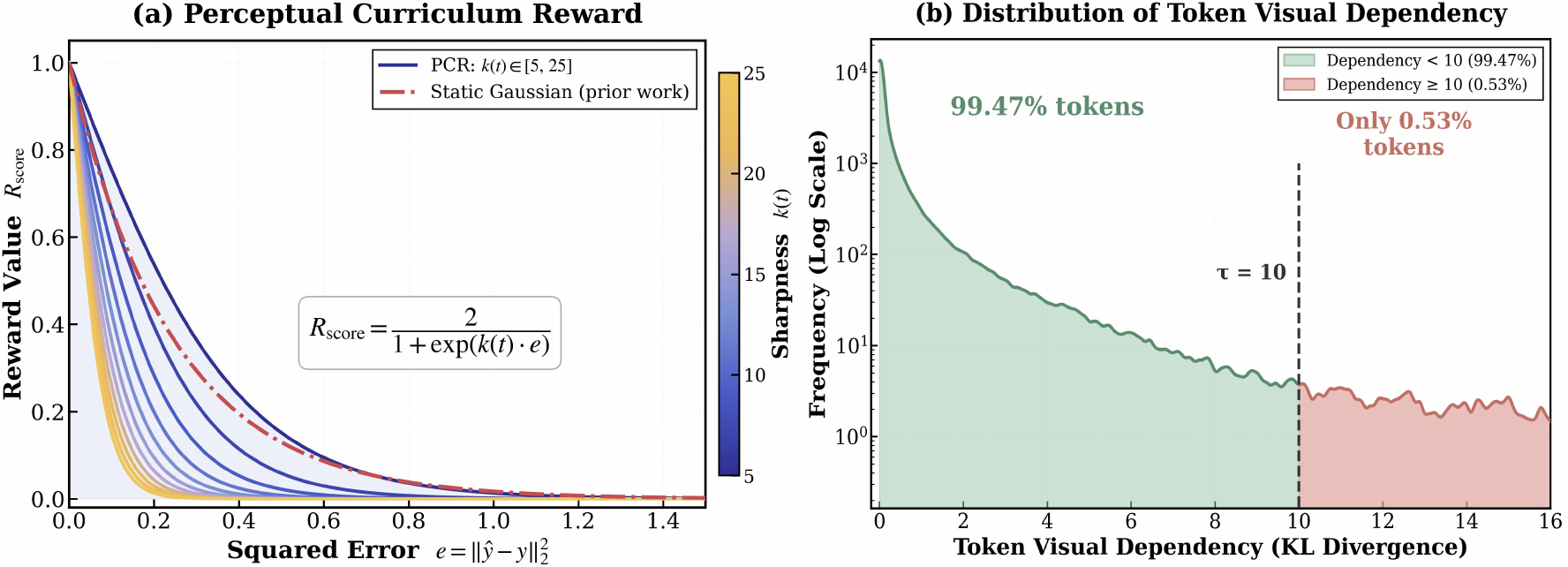}
  \caption{\textbf{Motivation for PCR and EGF.} (a)~Unlike static rewards (dashed) with fixed sensitivity, PCR dynamically increases sharpness $k(t)$ from broad exploration to fine-grained discrimination. (b)~Only 0.53\% of tokens exhibit high visual dependency ($\geq\!\tau$), confirming that uniform gradient allocation wastes signal on non-perceptual tokens.}
  \label{fig:motivation}
  \vspace{-10pt}
\end{figure}

Training such iMCoT trajectories with outcome-driven RL, however, is substantially more challenging than text-only CoT learning.
First, supervision is sparse and delayed: a final scalar outcome must guide many intermediate \emph{where-to-look} decisions, and traditional static score rewards~\cite{li2025q} can be either too harsh early on or too lenient later, causing reward saturation (Fig.~\ref{fig:motivation}a).
Second, credit assignment is harder in long iMCoT sequences that include tool-call syntax and observation tokens, since naively backpropagating the same outcome to all tokens dilutes the signal for the few visually grounded, decision-critical tokens. We quantify token-level visual dependency via KL divergence between predictions conditioned on the original and a distortion-perturbed image, and find that only $\sim$0.53\% of tokens are highly visually grounded (Fig.~\ref{fig:motivation}b). This confirms that broadcasting a uniform learning signal to all tokens dilutes gradients on the few diagnosis-critical ones.

To address these challenges, we introduce two complementary techniques:
\textbf{(1) Perceptual Curriculum Reward (PCR)}: a coarse-to-fine reward curriculum that starts with a loose tolerance to encourage early exploration, then progressively tightens it to demand finer quality discrimination and avoid late-stage saturation.
\textbf{(2) Evidence Gradient Filtering (EGF)}: focuses optimization on visually dependent, diagnosis-critical tokens (e.g., locating degraded regions and identifying artifacts), while suppressing gradients from tool-call syntax and other redundant tokens, thereby reducing noise and improving credit assignment in long trajectories.
Beyond producing grounded assessments, Q-DeepSight also provides a natural pathway toward \emph{image enhancement}.
Leveraging its localized and explainable quality diagnoses, we further explore \textbf{(3) Perceptual-in-Generation (PiG)}, an iMCoT-inspired reasoning-and-editing framework that injects explicit perceptual awareness into image synthesis.
The key idea is to move from an ``assess-only'' paradigm to an ``assess-and-refine'' loop, where \emph{image editing} serves as the bridge that converts quality understanding into actionable generative guidance.
Concretely, PiG first produces an initial image (or restoration output) from a prompt/input, then invokes Q-DeepSight to identify failure modes with grounded evidence (i.e., \emph{where} the quality degrades and \emph{why}), and finally translates these diagnoses into targeted edit instructions (e.g., region-specific refinement) to update the image. This diagnose--edit cycle can be repeated for multiple rounds to progressively improve the output. This closes the loop between perception and generation, showcasing a perceptual critic that assesses quality and guides targeted refinements.
Our contributions are as follows:
\begin{figure*}[t!]
  \centering
  \includegraphics[width=0.97\textwidth]{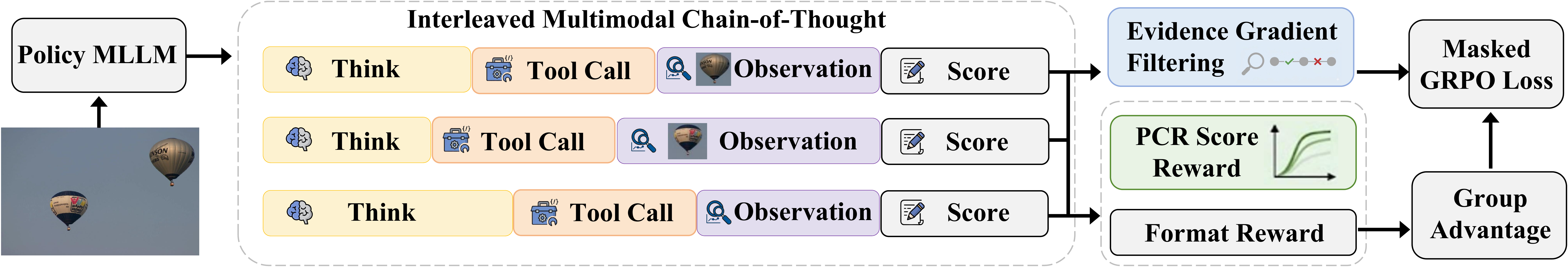}
  \caption{\textbf{Training pipeline of Q-DeepSight.} The policy MLLM generates multiple iMCoT rollouts, each interleaving Think, Tool Call, Observation, and Score steps. PCR (Sec.~\ref{sec:pcr}) provides a coarse-to-fine score reward via dynamic sharpness scheduling, while EGF (Sec.~\ref{sec:egf}) filters policy gradients to focus on visually grounded tokens identified by KL-based dependency measurement. }
  \label{fig:pipeline}
  \vspace{-10pt}
\end{figure*}
\begin{itemize}
    \item We reformulate IQA from a passive, single-look metric into an active, evidence-seeking sequential decision process. To this end, we propose \emph{\textbf{Q-DeepSight}}, a ``think-with-image'' framework that equips iMCoT with tool-augmented evidence acquisition, enabling iterative, localized, and strictly evidence-backed quality diagnoses.
    \item To mitigate the reward sparsity and credit assignment bottlenecks inherent in RL-based iMCoT training for IQA, we introduce Perceptual Curriculum Reward (PCR) and Evidence Gradient Filtering (EGF). These techniques jointly facilitate robust convergence and explicitly concentrate policy gradients on the perception of visual distortions.
    \item We further explore {Perceptual-in-Generation (PiG)}, a training-free assess-and-refine pipeline that translates Q-DeepSight's localized diagnoses into targeted editing instructions for iterative image improvement, closing the loop between quality assessment and visual enhancement.
\end{itemize}

%% file: sec/2_related_work.tex
\section{Related Work}
\subsection{MLLM-based IQA: Scores and Descriptions}
Traditional IQA is typically formulated as predicting a single perceptual quality score (or ranking) aligned with human opinion, using either full-reference similarity measures or no-reference regression models~\cite{wang2004image,mittal2012no,BRISQUE}. Recent MLLM-based approaches exploit strong vision--language representations for score regression (e.g., Q-Align~\cite{wu2023q} and DeQA-Score~\cite{you2025teaching}), but often provide limited grounded evidence for \emph{where} quality degrades and \emph{why}. This score-only formulation is often insufficient when IQA serves as an in-the-loop critic, where actionable, localized evidence is required for downstream refinement. In contrast, description-centric methods aim to generate natural-language critiques that improve interpretability, typically via supervised fine-tuning on curated rationales~\cite{descriptqa}. To reduce annotation cost, outcome-driven RL has been adopted to encourage quality-aware rationales using only a scalar quality objective~\cite{liu2023visual,wu2025visualquality,liang2026zoom}.
Despite improved rationales, most existing methods still follow a \emph{single-look, language-only} pipeline: after global image encoding, the model reasons primarily in text, making it hard to validate sparse, region-dependent artifacts or adaptively re-check visual evidence during reasoning.

\subsection{Interleaved Reasoning for Visual Grounding}
The limitations of ``language-only'' reasoning are not unique to IQA. For tasks that depend on fine-grained visual details, purely textual deliberation can be weakly grounded. A growing line of work therefore explores \textbf{interleaved image--text reasoning}, especially in Visual Question Answering (VQA)~\cite{hong2025deepeyesv2,su2025openthinkimg}, where models re-engage with the image during reasoning via tool use, such as iterative zooming or region selection~\cite{deepeyes}. This perception--action loop enables targeted evidence acquisition and improves both performance and interpretability~\cite{su2025thinking}.
Most existing interactive agents, however, rely on substantial supervised signals, including step-wise region annotations or human-authored reasoning trajectories~\cite{lai2025mini,wang2025pixel}. Such fine-grained supervision is costly and largely unavailable for IQA, limiting the applicability of interleaved reasoning to quality assessment.
Our work builds on iMCoT but targets IQA, where continuous perceptual scores and longer tool-augmented trajectories pose distinct challenges. We propose PCR to address reward sparsity and EGF to improve credit assignment for visually grounded tokens. We further introduce PiG, a training-free assess-and-refine pipeline that translates localized quality diagnoses into targeted editing instructions for iterative image improvement, bridging quality understanding and generation within a unified perceptual loop.

%% file: sec/3_methodology.tex
\section{Methodology}
\subsection{Overview}
We train a multimodal large language model (MLLM) for IQA that delivers accurate scores and faithful, evidence-grounded reasoning via an interleaved Multimodal Chain-of-Thought (iMCoT) process---alternating between textual analysis and tool-augmented visual evidence acquisition. 
Training such iMCoT trajectories with outcome-driven reinforcement learning (RL) presents key challenges of reward sparsity and poor credit assignment.
To address these challenges, we introduce two techniques.
\textbf{(1) Perceptual Curriculum Reward (PCR)}: A coarse-to-fine reward curriculum with loose early tolerance for exploration, progressive tightening for fine-grained quality discrimination, and late-stage saturation avoidance.
\textbf{(2) Evidence Gradient Filtering (EGF)}: Focuses optimization on visually dependent, diagnosis-critical tokens (e.g., degraded region localization, artifact identification) and suppresses gradients from tool-call syntax and redundant tokens, as shown in Fig.~\ref{fig:pipeline}.
Beyond producing grounded assessments, we further explore \textbf{(3) Perceptual-in-Generation (PiG)}, a training-free assess-and-refine application where Q-DeepSight's grounded diagnoses are translated into targeted edit instructions to iteratively improve generation outputs, as shown in Fig.~\ref{pig}.

\subsection{Think-with-Image via iMCoT}
\noindent\textbf{Rollout Formulation.}
We formalize Q-DeepSight's tool-augmented IQA reasoning as a Markov Decision Process (MDP) with external observations, building on prior work in tool-augmented interleaved reasoning~\cite{wang2025pixel,deepeyes}.
This formulation enables the model to alternate between textual reasoning and explicit visual evidence acquisition for accurate IQA: given $I_0$ and an IQA prompt, it generates a multi-turn iMCoT trajectory and may invoke a bbox-guided crop operator to zoom in on high-resolution local visual evidence.
Cropped regions are appended as additional observations, allowing subsequent steps to condition on the global image and accumulated local evidence; we limit tool invocations to 3 in our implementation.
The MDP state is the full iMCoT history up to step $t$:
\begin{equation*}
\hspace*{-0.7em}
  s_t = \{(X_0,I_0),(X_1,I_1),\ldots,(X_t,I_t)\} = \{X_{\le t}; I_{\le t}\}.
\end{equation*}
Here, $X_{\le t}$ is the accumulated text token sequence up to step $t$, and $I_{\le t}$ the accumulated image observation tokens (global image + cropped regions); VLM-non-generated special tokens are omitted for simplicity.
Conditioned on $s_t$, action token $a_t\sim\pi_\theta(\cdot\mid s_t)$ is sampled from policy $\pi_\theta$, being either a subsequent text token or a bbox tool call.

\noindent\textbf{Optimization.}
We adopt Group Relative Policy Optimization (GRPO) to enable efficient reinforcement learning (RL) fine-tuning~\cite{grpo}. For multi-turn agent trajectories, we apply a token-wise loss mask to ignore loss on observation tokens not generated by the model.

\subsection{Perceptual Curriculum Reward}\label{sec:pcr}
\noindent\textbf{Motivation.}
Directly using static error signals~\cite{li2025q,cao2025vqathinker} as reward creates numerical instability: large errors early in training can lead to excessively large gradients, while small errors late in training can yield vanishing gradients and slow down precision refinement.
To address this, we convert raw error into a bounded, smooth reward and gradually harden its sensitivity over training.

\noindent\textbf{Score reward shaping.}
Let $y$ be the score and $\hat y$ be the model prediction, and error $e=\left\lVert \hat y - y \right\rVert_2^2$.
We map $e$ into a bounded reward via a sharpness-controlled activation:
\begin{equation}
  R_{\mathrm{score}} = \frac{2}{1+\exp\!\left(k(t)\cdot e\right)}.
  \label{eq:rscore_sigmoid}
\end{equation}
A small $k(t)$ yields a smooth reward surface with broad gradient support, while a large $k(t)$ emphasizes small errors to facilitate late-stage precision. We also incorporate this dynamic sharpness into a continuous Gaussian reward, adopting an exponential variant in our implementation (Tab.~\ref{tab:ablation_training}):
\begin{equation}
  R_{\mathrm{score}} = \exp\!\left(-k(t)\cdot e\right) + \epsilon,
  \label{eq:rscore_exp}
\end{equation}
where $\epsilon$ is a small constant for numerical stability.

\noindent\textbf{Dynamic sharpness scheduling.}
Instead of using a fixed $k$, we increase it over training with a sigmoid schedule:
\begin{equation}
  k(t) = k_{\min} + (k_{\max}-k_{\min})\,\sigma\!\left(s\left(\frac{t}{T}-\tau\right)\right),
  \label{eq:sharpness}
\end{equation}
where $t$ is the current step, $T$ is the total number of steps, $\tau$ sets the transition center, and $s$ controls the steepness.
This coarse-to-fine curriculum produces dense feedback early and progressively hardens reward to prevent saturation later.

\noindent\textbf{Total reward.}
We optionally combine the score reward with a formatting reward $r_{\mathrm{fmt}}$~\cite{wang2025pixel,deepeyes} that encourages tool invocation, denoted as:
\begin{equation}
  R = R_{\mathrm{score}} + \lambda\, r_{\mathrm{fmt}},
\end{equation}
where $\lambda$ denotes a non-negative weighting hyperparameter that balances the contributions of the two reward terms.

\subsection{Evidence Gradient Filtering}\label{sec:egf}
\noindent\textbf{Token visual dependency.}
Multi-turn trajectories contain many tokens, yet usually only a subset is tightly grounded in visual evidence.
We quantify token visual dependency as the information gain provided by the image.
Given the state $s_t$, the true image context $I$ and a perturbed image $I'$, we define the token-level dependency score as
\begin{equation}
  S(s_t,I) := D_{\mathrm{KL}}\!\left(\pi_\theta(\cdot\mid s_t,I)\;\|\;\pi_\theta(\cdot\mid s_t,I')\right).
  \label{eq:vd}
\end{equation}
In contrast to common masking-based perturbations~\cite{wang2025perception}, we construct $I'$ via \textbf{distortion-based data augmentation} (agnostic to fine-grained quality evidence) by following the augmentation strategy in~\cite{reiqa}.
Intuitively, a substantial change in token prediction when the image is subject to distortion perturbation indicates that the token is likely relevant to quality prediction.

\noindent\textbf{Filtering policy gradients.}
For each trajectory $\tau_i$, we compute $S$ for all generated tokens and select the top-$k\%$ tokens as pivotal tokens, denoted by index set $\mathcal{K}_i$.
We define a binary mask $m_{i,t}=1$ if $t\in\mathcal{K}_i$ and $m_{i,t}=0$ otherwise.
During GRPO optimization, we weight the token-level objective by $m_{i,t}$ and always mask out observation tokens returned by crop/zoom.
Concretely, the masked objective is
\begin{equation}
\begin{aligned}
\mathcal{L}(\theta)=\mathbb{E}\Bigg[ & \frac{1}{G}\sum_{i=1}^{G}\frac{1}{|\mathcal{O}_i|}\sum_{t=1}^{|\mathcal{O}_i|} m_{i,t}\cdot \min\Big( r_{i,t}(\theta)\,\hat A'_i,\\
& \mathrm{clip}(r_{i,t}(\theta),1-\varepsilon,1+\varepsilon)\,\hat A'_i \Big)\Bigg]
\\
&\; -\; \beta_{\mathrm{KL}}\, D_{\mathrm{KL}}\!\left(\pi_\theta\;\|\;\pi_{\mathrm{ref}}\right).
\end{aligned}
\label{eq:grpo_masked}
\end{equation}
where $G$ is the group size, $\mathcal{O}_i$ denotes the generated tokens in trajectory $i$ (excluding observations), $r_{i,t}(\theta)$ is the importance ratio, $\hat A'_i$ is the group-relative advantage, and $\beta_{\mathrm{KL}}$ controls the KL regularization to a reference policy $\pi_{\mathrm{ref}}$.
This filtering mitigates gradient noise from generic tokens, directing learning toward quality-predictive visual tokens.
\begin{figure}[t]
  \centering
  \includegraphics[width=0.47\textwidth]{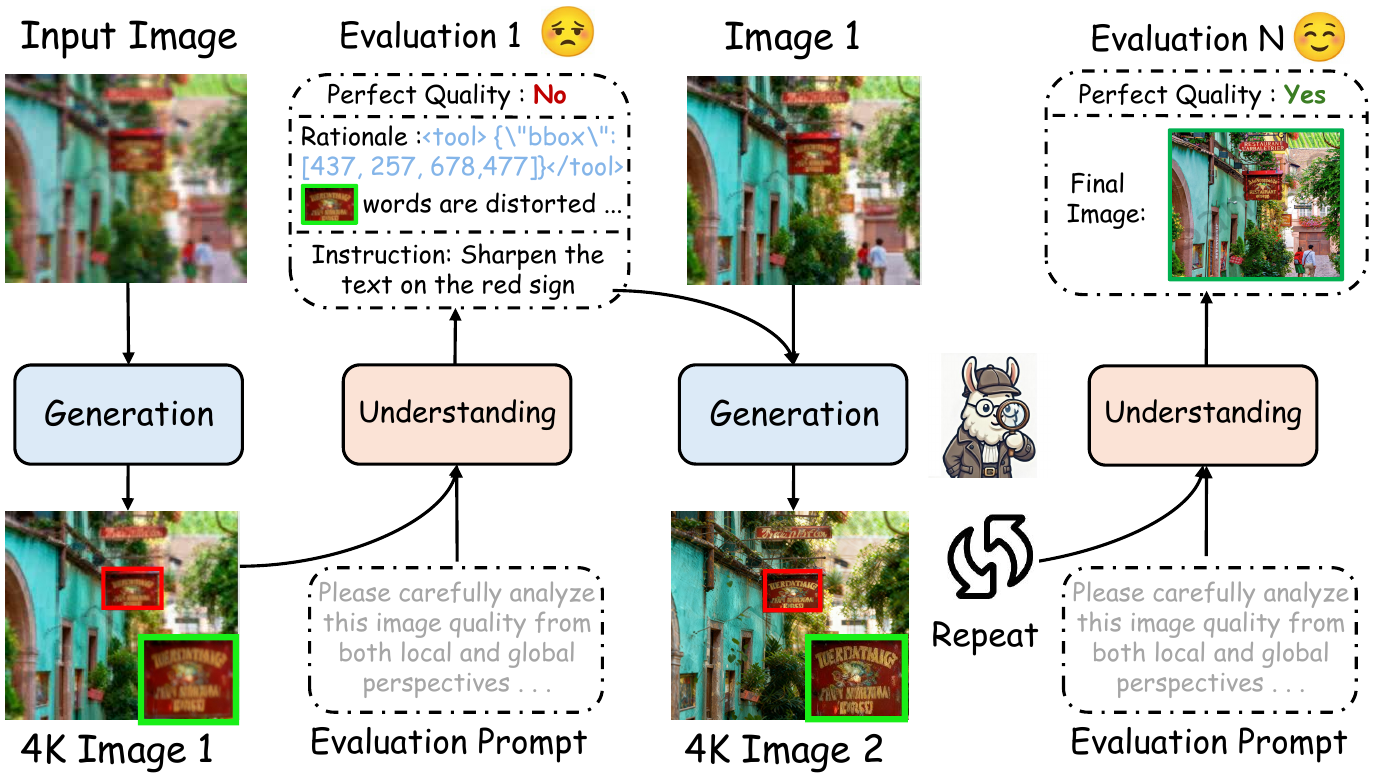}
  \vspace{-5pt}
  \caption{Overview of \textbf{Perceptual-in-Generation}. Given a mixed-degradation image, Q-DeepSight diagnoses quality issues with grounded evidence (where and why), then generates targeted edit instructions for iterative refinement.}
  \label{pig}
  \vspace{-10pt}
\end{figure}
\begin{table*}[t!]
\centering
\caption{\textbf{PLCC / SRCC comparison on the score regression tasks between our method and other competitive IQA methods.} All methods except handcrafted ones are trained on the \textbf{KonIQ dataset}. Q-Insight$^{\dagger}$ and Q-DeepSight$^{\dagger}$ denote models trained on a combined dataset (KonIQ + DQ-7K for distortion classification), following Q-Insight~\cite{li2025q}'s default setting.}
\vspace{-10pt}
\label{tab:main_results}
\setlength{\tabcolsep}{4pt}
\renewcommand{\arraystretch}{1.15}
\resizebox{\textwidth}{!}{
\begin{tabular}{llccccccc}
\toprule
\textbf{Category} & \textbf{Methods} & \textbf{KonIQ} & \textbf{SPAQ} & \textbf{KADID} & \textbf{PIPAL} & \textbf{LiveW} & \textbf{AGIQA} & \textbf{CSIQ} \\
\midrule
\midrule
\multirow{5}{*}{Non-MLLM} 
& HyperIQA~\cite{hypernet} & 0.917 / 0.906 & 0.791 / 0.788 & 0.506 / 0.468 & 0.410 / 0.403 & 0.772 / 0.749 & 0.702 / 0.640 & 0.752 / 0.717 \\
& DBCNN~\cite{pan2022dacnn} & 0.884 / 0.875 & 0.812 / 0.806 & 0.497 / 0.484 & 0.384 / 0.381 & 0.773 / 0.755 & 0.730 / 0.641 & 0.586 / 0.572 \\
& MUSIQ~\cite{ke2021musiq} & 0.924 / 0.929 & 0.868 / 0.863 & 0.575 / 0.556 & 0.431 / 0.431 & 0.789 / 0.830 & 0.722 / 0.630 & 0.771 / 0.710 \\
& ManIQA~\cite{yang2022maniqa} & 0.849 / 0.834 & 0.768 / 0.758 & 0.499 / 0.465 & 0.457 / 0.452 & 0.849 / 0.832 & 0.723 / 0.636 & 0.623 / 0.627 \\
\midrule
\multirow{4}{*}{\shortstack{MLLM \\ (w/o reasoning)}}
& CLIP-IQA+~\cite{wang2023exploring} & 0.909 / 0.895 & 0.866 / 0.864 & 0.653 / 0.654 & 0.427 / 0.419 & 0.832 / 0.805 & 0.736 / 0.685 & 0.772 / 0.719 \\
& C2Score~\cite{zhu2024adaptive} & 0.923 / 0.910 & 0.867 / 0.860 & 0.500 / 0.453 & 0.354 / 0.342 & 0.786 / 0.772 & 0.777 / 0.671 & 0.735 / 0.705 \\
& Q-Align~\cite{wu2023q} & 0.941 / 0.940 & 0.886 / 0.887 & 0.674 / 0.684 & 0.403 / 0.419 & 0.853 / 0.860 & 0.772 / 0.735 & 0.671 / 0.737 \\
& DeQA-Score~\cite{you2025teaching} & 0.953 / 0.941 & 0.895 / 0.896 & 0.694 / 0.687 & 0.472 / 0.478 & 0.892 / 0.879 & 0.809 / 0.729 & 0.787 / 0.744 \\
\midrule
\multirow{7}{*}{\shortstack{MLLM \\ (w/ reasoning)}}
& Qwen-SFT~\cite{bai2025qwen2} & 0.889 / 0.866 & 0.874 / 0.875 & 0.668 / 0.663 & 0.473 / 0.442 & 0.734 / 0.728 & 0.813 / 0.739 & 0.674 / 0.650 \\
& Q-Insight~\cite{li2025q} & 0.918 / 0.895 & 0.903 / 0.903 & 0.702 / 0.702 & 0.458 / 0.435 & 0.870 / 0.839 & 0.816 / 0.766 & 0.685 / 0.640 \\
& VisualQuality-R1~\cite{wu2025visualquality} & 0.910 / 0.896 & 0.889 / 0.892 & 0.703 / 0.712 & 0.451 / 0.441 & 0.856 / 0.827 & 0.817 / 0.760 & 0.768 / 0.707 \\
& Zoom-IQA~\cite{liang2026zoom} & {0.938} / {0.922} & {0.902} / {0.900} & {0.701} / {0.700} & {0.468} / {0.465} & {0.887} / {0.870} & {0.816} / {0.765} & {0.797} / {0.754} \\
& \textbf{Q-DeepSight (Ours)} &  \textbf{0.953} / \textbf{0.942} & {0.905} / {0.906} & {0.722} / {0.748} & \textbf{0.510} / \textbf{0.502} & \textbf{0.914} / {0.891} & {0.815} / {0.757} & {0.843} / {0.806} \\
& Q-Insight$^{\dagger}$~\cite{li2025q} & 0.933 / 0.916 & 0.907 / 0.905 & 0.742 / 0.736 & 0.486 / 0.474 & 0.893 / 0.865 & 0.811 / 0.764 & 0.870 / 0.824 \\
\rowcolor{cyan!10}
& \textbf{Q-DeepSight$^{\dagger}$ (Ours)} & {0.949} / {0.936} & \textbf{0.912} / \textbf{0.911} & \textbf{0.754} / \textbf{0.772} & {0.501} / {0.496} & \textbf{0.914} / \textbf{0.893} & \textbf{0.822} / \textbf{0.768} & \textbf{0.893} / \textbf{0.847} \\
\bottomrule
\end{tabular}}
\vspace{-10pt}
\end{table*}
\subsection{Perceptual-in-Generation}
Beyond IQA benchmarking, we explore a \textbf{training-free} \emph{assess-and-refine} loop that turns grounded quality understanding into actionable generation/restoration improvements.
Given a user input $u$ (e.g., a mix-degraded image input with its condition), we first obtain an initial restored image $I_0$.
At each iteration, Q-DeepSight acts as a perceptual critic and performs iMCoT-style evidence acquisition (crop/zoom) to produce grounded diagnoses that localize \emph{where} the quality degrades and explain \emph{why}.
These diagnoses are then translated into targeted, potentially region-aware edit instructions, which are applied by an image editing module, as shown in Fig.~\ref{pig}.
We formalize PiG as:
\begin{equation}
I_{i-1} \xrightarrow{\mathrm{Diagnose}} (d_i, e_i) \xrightarrow{\mathrm{Edit}} I_i,\quad i=1,\ldots,K.
\end{equation}
where $d_i$ denotes grounded diagnoses (e.g., failure modes with supporting evidence) and $e_i$ denotes the corresponding edit instructions for quality refinement.
In practice, we conduct a maximum of K=3 refinement rounds and enable early stopping: the loop terminates once Q-DeepSight judges the output as satisfactory based on $d_i$, and we take the corresponding $I_i$ as the final output. If no early stop is triggered after $K$ iterations, $I_K$ is used as the final result.

%% file: sec/4_experiments.tex
\begin{table}[t!]
\centering
\caption{Performance of the \textbf{Multi-Image Quality Comparison Task} on the Super-resolution Benchmark SRbench. Reg-Acc and Gen-Acc denote the accuracy of regression-based and generation-based restoration methods, respectively.}
\label{tab:srbench}
\setlength{\tabcolsep}{4pt}
\renewcommand{\arraystretch}{1.11}
\resizebox{\columnwidth}{!}{
\begin{tabular}{lccc}
\toprule
\textbf{Method} & \textbf{Reg-Acc} & \textbf{Gen-Acc} & \textbf{Overall-Acc} \\
\midrule
\multicolumn{4}{l}{\textit{Score-Based}} \\
PSNR & 80.07\% & 41.70\% & 34.70\% \\
SSIM~\cite{wang2004image} & 83.00\% & 45.30\% & 37.40\% \\
LPIPS~\cite{LPIPS} & 82.00\% & 63.90\% & 65.80\% \\
DISTS~\cite{DISTS} & 83.30\% & 66.60\% & 72.40\% \\
AHIQ~\cite{lao2022attentions} & \underline{83.70\%} & 70.00\% & 68.40\% \\
TOPIQ~\cite{chen2024topiq} & \underline{83.70\%} & 63.90\% & 67.00\% \\
A-FINE~\cite{A-FINE} & 83.30\% & \textbf{78.90}\% & \textbf{82.40}\% \\
\midrule
\multicolumn{4}{l}{\textit{Description-Based (Zero-Shot)}} \\
DepictQA~\cite{you2024depicting} & 73.00\% & 61.64\% & 62.96\% \\
Q-Insight~\cite{li2025q} & 78.67\% & 68.64\% & 75.51\% \\
VisualQuality-R1~\cite{wu2025visualquality} & 77.05\% & 50.42\% & 59.62\% \\
\midrule
\rowcolor{cyan!10}
\textbf{Q-DeepSight (Ours)} & \textbf{88.53\%} & \underline{74.14\%} & \underline{79.11\%} \\
\bottomrule
\end{tabular}}
\end{table}

\begin{table}[t!]
\centering
\caption{\textbf{Performance on processing-related distortion datasets.} Comparison of SRCC and PLCC results across de-blurring~\cite{liu2013no}, super-resolution (SRIQA-Bench)~\cite{A-FINE}, and image generation datasets~(AIGC-3K)~\cite{li2023agiqa}.}
\label{tab:processing_distortion}
\setlength{\tabcolsep}{4pt}
\renewcommand{\arraystretch}{1.05}
\resizebox{\columnwidth}{!}{
\begin{tabular}{lcccc}
\toprule
\textbf{Method} & \textbf{Deblur} & \textbf{SRIQA} & \textbf{AIGC} & \textbf{Avg} \\
\midrule
\multicolumn{5}{l}{\textit{SRCC}} \\
LIQE~\cite{LIQE} & 0.797 & 0.743 & 0.653 & 0.731 \\
Q-Align~\cite{wu2023q} & 0.761 & 0.684 & 0.682 & 0.709 \\
DeQA-Score~\cite{you2025teaching} & 0.785 & 0.710 & 0.738 & 0.744 \\
Q-Insight~\cite{li2025q} & 0.831 & 0.724 & 0.749 & 0.768 \\
{VisualQuality-R1~\cite{wu2025visualquality}} & \underline{0.845} & \underline{0.752} & \underline{0.754} & \underline{0.784} \\
\rowcolor{cyan!10}
\textbf{Q-DeepSight (Ours)} & \textbf{0.891} & \textbf{0.773} & \textbf{0.768} & \textbf{0.811} \\
\midrule
\multicolumn{5}{l}{\textit{PLCC}} \\
LIQE~\cite{LIQE} & 0.712 & 0.775 & 0.653 & 0.713 \\
Q-Align~\cite{wu2023q} & 0.802 & 0.713 & 0.705 & 0.740 \\
DeQA-Score~\cite{you2025teaching} & 0.838 & 0.763 & 0.790 & 0.797 \\
Q-Insight~\cite{li2025q} & 0.857 & 0.798 & 0.810 & 0.822 \\
VisualQuality-R1~\cite{wu2025visualquality} & \underline{0.879} & \textbf{0.824} & \underline{0.820} & \underline{0.841} \\
\rowcolor{cyan!10}
\textbf{Q-DeepSight (Ours)} & \textbf{0.897} & \underline{0.808} & \textbf{0.822} & \textbf{0.842} \\
\bottomrule
\end{tabular}}
\end{table}
\section{Experiments}
\subsection{Experimental Settings}
We initialize {Qwen3-VL-8B}~\cite{Qwen3-VL} as our base model. Following Q-Insight~\cite{li2025q} and Zoom-IQA~\cite{liang2026zoom}, we train two variants: trained on only KonIQ dataset, and trained on the combined dataset (KonIQ + DQ-7K~\cite{descriptqa}) with an auxiliary distortion classification setting.
In the GRPO algorithm, the generation number $N$ is set to 8, and the KL loss coefficient $\beta$ is set to $1 \times 10^{-3}$. The model is trained for 10 epochs with a total batch size of 128. During rollout generation, we use temperature 1.0 and top-p 1.0 for sampling. We enable multi-turn agent interactions with a maximum of 3 turns. For PCR, we set \(k_{\min}=5\) and \(k_{\max}=25\). For EGF, we select the top-\(40\%\) tokens by visual dependency scores as pivotal tokens.

\noindent\textbf{IQA Datasets.}
We evaluate Q-DeepSight on a diverse suite of NR-IQA benchmarks that cover four categories: \textbf{in-the-wild} distortions including KonIQ-10k~\cite{hosu2020koniq}, SPAQ~\cite{spaq} and LIVE Challenge~\cite{LIVEC}, \textbf{synthetic} distortions including KADID-10K~\cite{lin2019kadid}, PIPAL~\cite{pipal} and CSIQ~\cite{csiq}, \textbf{AI-generated} imagery including AGIQA~\cite{li2023agiqa}, and \textbf{processing-related} distortions including de-blurring~\cite{liu2013no} and super-resolution (SRIQA-Bench)~\cite{A-FINE}.
We additionally include a high-resolution IQA benchmark UHD~\cite{uhdiqa} to evaluate model robustness under fine-grained details.
Unless noted otherwise, we follow the standard train/test splits of each dataset as described in~\cite{li2025q}.
Following prior IQA literature, we report {Spearman rank-order correlation coefficient (SRCC)} and {Pearson linear correlation coefficient (PLCC)} between predicted scores and human opinion scores.

\subsection{Image Quality Assessment}
\noindent\textbf{Performance on standard NR-IQA benchmarks.}
Table~\ref{tab:main_results} compares Q-DeepSight with state-of-the-art IQA methods across seven datasets.
Compared to the strongest MLLM baseline Q-Insight$^{\dagger}$, Q-DeepSight$^{\dagger}$ delivers consistent gains on all datasets, with the largest improvements reaching \textbf{+2.1} correlation points on LiveW and \textbf{+2.3} points on CSIQ.
These results highlight the benefit of tool-augmented iMCoT reasoning together with Perceptual Curriculum Reward (PCR).

\noindent\textbf{Restoration and AIGC quality assessment.}
Tables~\ref{tab:srbench} and \ref{tab:processing_distortion} further demonstrate strong generalization to restored images and AI-generated images.
On SRbench, Q-DeepSight is evaluated in a zero-shot setting and consistently outperforms Q-Insight, improving Reg-Acc, Gen-Acc, and Overall-Acc by \textbf{9.86}, \textbf{5.50}, and \textbf{3.60} points, respectively.
On processing-related distortion datasets (Table~\ref{tab:processing_distortion}), Q-DeepSight improves over VisualQuality-R1 on AIGC-3K by \textbf{1.4} points in SRCC and \textbf{0.2} points in PLCC, and achieves the best average correlation across most scenarios.
\begin{table}[t!]
\centering
\caption{\textbf{Performance comparison on DRealSR for super-resolution task.} Best results are in \textbf{bold} and second-best results are \underline{underlined}. $\uparrow$ / $\downarrow$ indicates higher/lower is better.}
\label{tab:drealsr}
\setlength{\tabcolsep}{1pt}
\renewcommand{\arraystretch}{1.25}
\resizebox{\columnwidth}{!}{
\begin{tabular}{lccccc}
\toprule
\textbf{Method} & \textbf{PSNR$\uparrow$} & \textbf{DISTS$\downarrow$} & \textbf{MUSIQ$\uparrow$} & \textbf{MANIQA$\uparrow$} & \textbf{TOPIQ$\uparrow$} \\
\midrule
DiffBIR~\cite{diffbir} & 26.08 & 0.2564 & 61.81 & 0.4612 & 0.6084 \\
SeeSR~\cite{seesr} & \underline{28.14} & 0.2241 & 55.89 & 0.3976 & 0.5436 \\
PASD~\cite{pasd} & \textbf{28.18} & \underline{0.2108} & 51.42 & 0.3595 & 0.4587 \\
ResShift~\cite{resshift} & 27.39 & 0.3077 & 40.58 & 0.2457 & 0.3414 \\
SinSR~\cite{sinsr} & 26.72 & 0.2624 & 53.36 & 0.3677 & 0.4959 \\
OSEDiff~\cite{osediff} & 25.60 & 0.2158 & \underline{65.24} & 0.4879 & \underline{0.6273} \\
S3Diff~\cite{s3diff} & 26.18 & \textbf{0.2099} & 63.34 & 0.4635 & 0.6181 \\
PURE~\cite{pure} & 23.04 & 0.2674 & 60.68 & 0.4362 & 0.5888 \\
UARE~\cite{uare} & 21.31 & 0.2613 & \textbf{67.71} & \underline{0.5121} & \textbf{0.6652} \\
\midrule
Bagel~\cite{bagel} & 25.21 & 0.2628 & 32.02 & 0.3717 & 0.2793 \\
\rowcolor{cyan!10}
\textbf{+ Q-DeepSight Prompt} & 20.80 & 0.2882 & 58.90 & \textbf{0.5535} & 0.5358 \\
\quad $\Delta$ & \textcolor{lightgray}{\textbf{-4.41}} & \textcolor{lightgray}{\textbf{+0.025}} & \textcolor{customgreen}{\textbf{+26.88}} & \textcolor{customgreen}{\textbf{+0.182}} & \textcolor{customgreen}{\textbf{+0.257}} \\
\bottomrule
\end{tabular}}
\end{table}

\subsection{Super Resolution and Image Enhancement}
\noindent\textbf{Task setup.}
Following~\cite{s3diff,uare}, we compare Q-DeepSight with prior methods on center-cropped DRealSR~\cite{drealsr} with $512 \times 512$ resolution, performing $4\times$ SR.
For DIV4K-50~\cite{4kagent}, we restore \(256 \times 256\) inputs directly to \(4\mathrm{K}\) high-resolution images, which requires preserving fine details and global consistency under large upscaling ratios.

\noindent\textbf{Two Perceptual-in-Generation strategies.}
We consider two ways to translate Q-DeepSight's grounded IQA understanding into actionable improvement signals: (1) one-shot \emph{prompt-based} guidance, and (2) \emph{iterative} critic feedback (Fig.~\ref{pig}).
We test strategy (1) on DRealSR because its resolution is relatively low and the task is simpler; the model can readily capture both global context and local cues, making one-shot guidance sufficient for enhancement.
We use strategy (2) on DIV4K-50~\cite{4kagent} because restoring to \(4\mathrm{K}\) is substantially more challenging and often requires localized, fine-grained corrections; iterative critic feedback enables progressive improvement of regional artifacts.

\begin{figure*}[t!]
  \centering
  \includegraphics[width=0.97\textwidth]{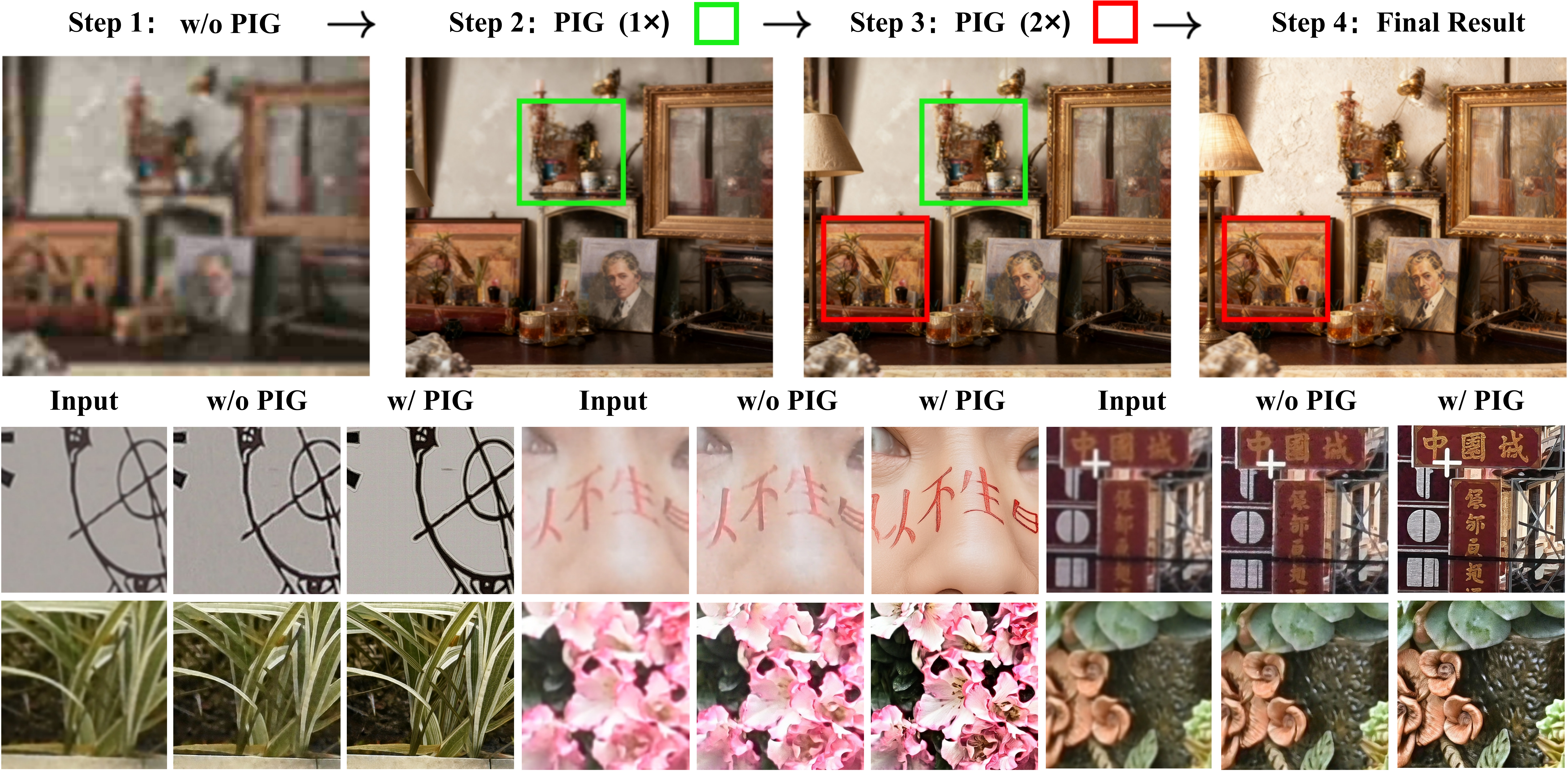}
  \caption{Qualitative visualization of \textbf{PiG}. The model alternates between textual reasoning and bbox-guided crop/zoom to acquire additional evidence, and provides grounded feedback for iterative refinement.}
  \label{visual}
  \vspace{-10pt}
\end{figure*}

\noindent\textbf{Results.}
Table~\ref{tab:drealsr} shows that Q-DeepSight's prompt-based guidance yields substantial perceptual gains for super-resolution on DRealSR.
Firstly, it increases MUSIQ by \textbf{26.88} points and improves TOPIQ by \textbf{0.257}.
Secondly, it improves MANIQA by \textbf{0.182}, suggesting better perceptual fidelity.
Note that the PSNR decrease is acceptable under the well-known perception-distortion tradeoff~\cite{blau2018perception}, where perceptual quality gains inherently come at the cost of pixel-level fidelity.
Table~\ref{tab:div4k50} indicates that iterative critic feedback consistently improves restoration quality on DIV4K-50: with two refinement rounds, NIQE decreases by \textbf{0.54}, while CLIPIQA and MUSIQ improve by \textbf{0.031} and \textbf{5.14}, respectively.
Qualitative results are provided in Fig.~\ref{visual}. Q-DeepSight accompanies restoration with fine-grained, language-aligned quality analysis, identifying failure modes such as illegible text, heavy pixelation, and ringing artifacts, and suggesting targeted refinements.
Moreover, the guided outputs exhibit clearer text and more regular local patterns, such as windows and brick textures, with markedly fewer artifacts.
\begin{table}[t!]
\centering
\caption{\textbf{Performance comparison on DIV4K-50 dataset for image restoration task.} Methods are evaluated at $16\times$ upsampling by default; $^{\dagger}$ denotes two-stage $4\times$ upsampling. }
\label{tab:div4k50}
\setlength{\tabcolsep}{1pt}
\renewcommand{\arraystretch}{1.1}
\resizebox{\columnwidth}{!}{
\begin{tabular}{lcccc}
\toprule
\textbf{Method} & \textbf{NIQE$\downarrow$} & \textbf{CLIPIQA$\uparrow$} & \textbf{MUSIQ$\uparrow$} & \textbf{MANIQA$\uparrow$} \\
\midrule
HAT-L$^{\dagger}$~\cite{hatl} & 11.86 & 0.4699 & 22.82 & 0.3270 \\
DiffBIR$^{\dagger}$~\cite{diffbir} & \underline{3.36} & \textbf{0.7588} & 37.17 & \textbf{0.5916} \\
DiffBIR~\cite{diffbir} & \textbf{2.65} & 0.7078 & 38.59 & \underline{0.5858} \\
OSEDiff$^{\dagger}$~\cite{osediff} & 4.88 & \underline{0.7201} & \textbf{39.88} & 0.5482 \\
OSEDiff~\cite{osediff} & 8.37 & 0.5680 & 25.07 & 0.4210 \\
PiSA-SR$^{\dagger}$~\cite{pisasr} & 5.01 & 0.7141 & 38.22 & 0.5364 \\
PiSA-SR~\cite{pisasr} & 9.30 & 0.5549 & 24.51 & 0.3861 \\
AgenticIR~\cite{agenticir} & 5.13 & 0.5614 & \underline{39.55} & 0.4814 \\
\midrule
Seedream~\cite{seedream2025seedream} & 5.80 & 0.3739 & 25.82 & 0.4551 \\
\rowcolor{cyan!10}
\textbf{+ PiG with 1 iteration} & 5.39 & 0.3788 & 28.06 & 0.4688 \\
\quad $\Delta$ & \textcolor{customgreen}{\textbf{-0.41}} & \textcolor{customgreen}{\textbf{+0.005}} & \textcolor{customgreen}{\textbf{+2.24}} & \textcolor{customgreen}{\textbf{+0.014}} \\
\rowcolor{cyan!10}
\textbf{+  PiG with 2 iterations} & 5.26 & 0.4053 & 30.96 & 0.4703 \\
\quad $\Delta$ & \textcolor{customgreen}{\textbf{-0.54}} & \textcolor{customgreen}{\textbf{+0.031}} & \textcolor{customgreen}{\textbf{+5.14}} & \textcolor{customgreen}{\textbf{+0.015}} \\
\bottomrule
\end{tabular}}
\end{table}
\begin{table*}[t]
\centering
\caption{{Ablation Study of PCR and EGF Methods.} Results are reported as PLCC / SRCC. \textbf{EGF+Gauss}: EGF with continuous Gaussian reward~\cite{liang2026zoom}. \textbf{EGF+Gauss+Rank}: EGF with Gaussian and ranking reward~\cite{wu2025visualquality}. \textbf{Binary}: Binary reward.}
\label{tab:ablation_training}
\setlength{\tabcolsep}{4pt}
\renewcommand{\arraystretch}{1.15}
\resizebox{\textwidth}{!}{
\begin{tabular}{lcccccccc}
\toprule
\textbf{Method} & \textbf{KonIQ} & \textbf{SPAQ} & \textbf{KADID} & \textbf{PIPAL} & \textbf{LiveW} & \textbf{AGIQA} & \textbf{CSIQ} & \textbf{UHD} \\
\midrule
Binary & 0.947 / 0.932 & 0.902 / 0.905 & 0.706 / 0.737 & 0.489 / 0.487 & 0.910 / 0.886 & 0.811 / 0.756 & 0.829 / 0.792 &  0.674 / 0.613 \\
EGF + Binary & 0.947 / 0.936 & 0.901 / 0.904 & 0.710 / 0.744 & 0.491 / 0.483 & 0.905 / 0.877 & 0.812 / 0.762 & 0.843 / 0.802 & 0.665 / 0.601 \\
EGF + Gauss & 0.951 / 0.940 & \textbf{0.908} / \textbf{0.909} & 0.714 / 0.728 & 0.502 / 0.494 & 0.907 / 0.882 & 0.800 / 0.735 & 0.815 / 0.785 &  0.686 / 0.634 \\
EGF + Gauss + Rank & 0.952 / 0.943 & 0.901 / 0.902 & 0.710 / 0.723 & 0.505 / 0.495 & 0.916 / 0.894 & 0.798 / 0.743 & 0.823 / 0.792 & 0.689 / 0.618 \\
\rowcolor{cyan!10}
EGF + PCR (Ours) & \textbf{0.953} / \textbf{0.942} & {0.905} / {0.906} & \textbf{0.722} / \textbf{0.748} & \textbf{0.510} / \textbf{0.502} & \textbf{0.914} / \textbf{0.891} & \textbf{0.815} / \textbf{0.757} & \textbf{0.843} / \textbf{0.806} & \textbf{0.703} / \textbf{0.640} \\
\bottomrule
\end{tabular}}
\vspace{-5pt}
\end{table*}

\begin{table}[t!]
\centering
\caption{{Ablation study on reasoning strategies.} Results are reported as PLCC / SRCC. \textbf{Text-only CoT}: standard text-based chain-of-thought reasoning. \textbf{iMCoT}: our multimodal chain-of-thought reasoning with tools. \textbf{Autothink CoT}: automated reasoning path selection. \textbf{$\dagger$}: model ensemble. $\dagger_{\text{avg}}$ denotes score averaging; $\dagger_{\text{best}}$ uses the prediction closest to GT.}
\label{tab:ablation}
\setlength{\tabcolsep}{2pt}
\renewcommand{\arraystretch}{1.25}
\resizebox{\columnwidth}{!}{
\begin{tabular}{lcccc}
\toprule
\textbf{Method} & \textbf{KonIQ} & \textbf{KADID} & \textbf{LiveW} & \textbf{CSIQ} \\
\midrule
Text-only CoT & 0.939 / 0.926 & \underline{0.768} / \underline{0.792} & 0.900 / 0.877 & 0.882 / 0.846 \\
Autothink CoT & \underline{0.947} / \underline{0.935} & 0.739 / 0.752 & \underline{0.904} / 0.884 & \textbf{0.894} / \textbf{0.847} \\
\rowcolor{cyan!10}
iMCoT (Ours) & \textbf{0.949} / \textbf{0.936} & \textbf{0.754} / \textbf{0.772} & \textbf{0.914} / \textbf{0.893} & \underline{0.893} / \textbf{0.847} \\
\midrule
\multicolumn{5}{c}{\textit{Ensemble Methods}} \\
Text $\dagger_{\text{avg}}$ iMCoT & 0.947 / 0.933 & 0.765 / 0.785 & 0.911 / 0.888 & 0.892 / 0.850 \\
iMCoT $\dagger_{\text{best}}$ iMCoT & 0.952 / 0.940 & 0.758 / 0.777 & 0.918 / 0.898 & 0.895 / 0.850 \\
Text $\dagger_{\text{best}}$ iMCoT & \textbf{0.967} / \textbf{0.964} & \textbf{0.808} / \textbf{0.842} & \textbf{0.939} / \textbf{0.928} & \textbf{0.908} / \textbf{0.869} \\
\bottomrule
\end{tabular}}
\end{table}
\noindent\textbf{Qualitative analysis of PiG.}
Fig.~\ref{visual} demonstrates PiG's progressive refinement across diverse scenarios. In the top row, the indoor scene shows clear artifact reduction: the green-boxed region transitions from severe ghosting to well-defined boundaries, and the mirror next to the green box gains sharpness with each iteration. The bottom grid showcases fine-grained improvements in detail, such as sharper curves in geometric clock patterns, legible rendering of Chinese characters, and architectural elements like door panels shifting from blurred to crisp. Additionally, natural textures, including leaves, show markedly improved definition.

\subsection{Ablation Studies}
\noindent\textbf{Effect of reasoning strategies.}
Table~\ref{tab:ablation} is designed to answer a practical question: \emph{does IQA benefit from think-with-image reasoning, or is text-only reasoning sufficient?}
Among single models, iMCoT achieves the best average performance, suggesting that explicit evidence acquisition improves IQA robustness in general.
At the same time, the gap narrows on datasets dominated by global distortions (e.g., KADID), where a coarse global impression can be sufficient and text-only reasoning remains competitive.
In contrast, when degradations are localized, iMCoT yields clearer gains, consistent with the need to actively inspect and ground judgments on regional evidence.
Ensembling further supports this view by revealing complementary failure modes.
The Text $\dagger_{\text{best}}$ iMCoT ensemble improves over single iMCoT by \textbf{2.2} points in PLCC and \textbf{4.7} points in SRCC, indicating that combining global heuristic reasoning with evidence-grounded inspection provides the most reliable assessments, which is a promising direction.

\noindent\textbf{Effect of EGF.} To verify EGF effectiveness, we conduct controlled ablations by pairing EGF with different reward designs. As shown in Table~\ref{tab:ablation_training}, EGF + Binary exhibits inconsistent performance across datasets, with improvements on some benchmarks but degradation on others, particularly on high-resolution UHD. This indicates that token-level filtering requires dense supervision signals---sparse binary rewards cannot effectively guide gradient allocation to perceptually-dependent tokens. In contrast, EGF + Gauss substantially improves performance across most datasets, especially on KADID and UHD. This confirms that EGF's gradient filtering mechanism relies on continuous, fine-grained rewards to properly assign credit and mitigate noise in high-resolution IQA scenarios.

\noindent\textbf{Effect of PCR.} To further validate Perceptual Curriculum Reward (PCR), we compared it to two common static reward designs in RL-based IQA: continuous Gaussian rewards~\cite{liang2026zoom} and ranking rewards~\cite{wu2025visualquality}. With EGF fixed, Table~\ref{tab:ablation_training} shows adding a ranking term to Gaussian reward (EGF + Gauss + Rank) provided limited benefit over Gaussian alone. In contrast, replacing static rewards with PCR achieved the best overall performance. For example, compared to EGF + Gauss + Rank, EGF + PCR improved UHD by \textbf{0.014} in PLCC and \textbf{0.022} in SRCC, and KADID by \textbf{0.025} in SRCC. This demonstrates PCR delivers a more effective coarse-to-fine supervision signal than static reward shaping, complementing EGF and better scaling to high-resolution quality assessment.
\begin{table}[t!]
\centering
\caption{\textbf{Visual evidence grounding analysis.} $\text{Acc}_{loc}$ measures the fraction of annotated pseudo-GT regions covered by Q-DeepSight's crops. Evidence perturbation replaces tool-returned crops with distorted versions.}
\label{tab:grounding}
\setlength{\tabcolsep}{3pt}
\renewcommand{\arraystretch}{1.15}
\resizebox{\columnwidth}{!}{
\begin{tabular}{lcccccc}
\toprule
& \multicolumn{3}{c}{\textbf{KonIQ}} & \multicolumn{3}{c}{\textbf{LiveW}} \\
\cmidrule(lr){2-4} \cmidrule(lr){5-7}
\textbf{Setting} & \textbf{PLCC} & \textbf{SRCC} & \textbf{ACC$_{loc}$} & \textbf{PLCC} & \textbf{SRCC} & \textbf{ACC$_{loc}$} \\
\midrule
\rowcolor{cyan!10}
\textbf{Q-DeepSight} & 0.953 & 0.942 & 69.4\% & 0.914 & 0.891 & 60.3\% \\
+ Evidence perturb. & 0.941 & 0.927 & - & 0.898 & 0.879 & - \\
\bottomrule
\end{tabular}}
\end{table}
\subsection{Validating Visual Evidence Grounding}
To investigate whether Q-DeepSight genuinely relies on the visual evidence from tool calls or simply predicts scores from global features with superficial reasoning traces, we conduct two complementary analyses: \emph{localization accuracy} and \emph{evidence perturbation}.

\noindent\textbf{Localization accuracy.}
Since current IQA datasets lack fine-grained annotations for distortion-relevant regions, we leverage Qwen3-VL-235B~\cite{Qwen3-VL} to generate pseudo-ground truth (pseudo-GT) bounding boxes as reference ``evidence regions''. Formally, let $R$ denote the predicted rationale region and $G$ the pseudo-GT bounding box. We define $\text{Acc}_{loc}$ as the fraction of the ground-truth area covered by the invoked region. As shown in Table~\ref{tab:grounding}, Q-DeepSight achieves a high $\text{Acc}_{loc}$ of 69.4\% on KonIQ and 60.3\% on LiveW, indicating that, even without process supervision, the model learns to localize distortion-relevant regions through outcome-driven RL alone.

\noindent\textbf{Evidence perturbation.}
To verify the functional necessity of iMCoT and test whether the model's predictions depend on the acquired visual evidence, we intentionally apply distortion perturbation to all tool-returned crop images during inference, while keeping the original input image unchanged.
If the model merely ignores the cropped evidence, this perturbation should have a negligible effect.
However, Table~\ref{tab:grounding} demonstrates a consistent performance drop under perturbation: SRCC decreases by \textbf{0.015} and \textbf{0.012} on KonIQ and LiveW, respectively. This confirms that Q-DeepSight actively conditions its assessment on acquired local evidence, validating the functional necessity of our iMCoT tool-calling framework.
\begin{table}[t!]
\centering
\caption{\textbf{Inference efficiency comparison} on LIVE Challenge. All models are evaluated on a single GPU with 90\,GB memory. ``Extra Image Tokens'' and ``Extra Latency'' denote the overhead introduced by multi-turn iMCoT reasoning.}
\vspace{-5pt}
\label{tab:efficiency}
\setlength{\tabcolsep}{1.5pt}
\renewcommand{\arraystretch}{1.1}
\resizebox{\columnwidth}{!}{
\begin{tabular}{lcccc}
\toprule
\textbf{Method} & \textbf{Total Tokens (t)} & \textbf{Latency (s)} & \textbf{Extra Image (t)}\\
\midrule
Q-Insight & 100.52 & 0.692 & 0.0\\
VisualQuality-R1 & 104.70 & 0.716 & 0.0\\
\midrule
Ours w/o EGF & 315.95 & 2.128 & 132.06\\
\rowcolor{cyan!10}
\textbf{Ours} & 177.51 & 1.361 & 75.19\\
\bottomrule
\end{tabular}}
\end{table}
\subsection{Computational Analysis}
Table~\ref{tab:efficiency} compares inference efficiency on the LiveW.
Unlike the single-turn method Q-Insight and VisualQuality-R1 that score an image in one forward pass, Q-DeepSight performs multi-turn iMCoT reasoning, trading higher latency for active visual evidence acquisition as a form of \emph{test-time scaling}: the model invests extra computation to actively gather visual evidence, which is essential for fine-grained, localized quality diagnosis that single-pass methods cannot achieve.
In practice, the overhead can be further reduced via KV caching and encode-prefill-decode disaggregation in modern serving infrastructure.
Notably, EGF plays a dual role: beyond improving credit assignment, it also reduces inference cost. Without EGF, the model generates substantially more tokens and invokes more tool calls, leading to 1.56$\times$ higher latency. With EGF, the model focuses on distortion-relevant tokens and avoids producing redundant reasoning, yielding more concise yet accurate trajectories.

%% file: sec/5_conclusion.tex
\section{Conclusion}
We proposed \textbf{Q-DeepSight}, a \emph{think-with-image} IQA framework that alternates between reasoning and tool-based evidence acquisition, producing grounded diagnoses of \emph{where} and \emph{why} quality degrades. 
To enable outcome-driven RL for tool-augmented trajectories, we introduced {Perceptual Curriculum Reward (PCR)} and {Evidence Gradient Filtering (EGF)} to alleviate reward sparsity and improve credit assignment.
Q-DeepSight achieves SOTA results across natural images, restoration outputs, and AIGC content while providing actionable feedback.
We also explored {Perceptual-in-Generation (PiG)} to demonstrate how quality understanding can guide iterative refinement.